\RequirePackage{snapshot}
\documentclass{article}

\usepackage[final]{bdl_2018}

\usepackage[utf8]{inputenc} %
\usepackage[T1]{fontenc}    %
\usepackage{hyperref}       %
\usepackage{url}            %
\usepackage{booktabs}       %
\usepackage{amsfonts}       %
\usepackage{nicefrac}       %
\usepackage{microtype}      %
\usepackage{graphicx}
\usepackage{amsmath, amssymb}
\usepackage{cleveref}

\title{$\beta$-VAEs can retain label information even at high compression}

\author{
	Emily Fertig \\
	Google Research\\
	\texttt{emilyaf@google.com}\\
	\And
	Aryan Arbabi\thanks{Work done while an intern at Google.}\\
	Vector Institute, University of Toronto\\
	\texttt{arbabi@cs.toronto.edu}\\
	\And
	Alexander. A. Alemi\\
	Google Research\\
	\texttt{alemi@google.com}
}

\begin{document}
	\maketitle

	\begin{abstract}
In this paper, we investigate the degree to which the encoding of a
$\beta$-VAE captures label information across multiple architectures
on Binary Static MNIST and Omniglot. 
Even though they are trained in a completely unsupervised manner,
we demonstrate that a $\beta$-VAE can retain a large amount
of label information, even when asked to learn a highly 
compressed representation.
\end{abstract}
}%

	\section{Introduction}

Our research question can be motivated by the following analogy:
imagine a bag filled with 100 balls, only 3 of which are red. Imagine
your friend reaches into the bag and ``randomly'' pulls out 70 balls.
If they happened to pull out all 3 of the red balls, you would not be
all that surprised (the probability of such an event is 34\%, the odds
are 2 to 1 against). If they reached in and ``randomly'' pulled out
only 10 balls, but managed to pull out all 3 of the red balls, you
should expect your friend cheated and was aware of some property of the
balls related to color (the probability of drawing all 3 red balls is 0.1\%, or
odds of 979 to 1). Do VAEs behave more like the first case or the second?

Variational autoencoders (VAEs) learn
compressed representations of their input, akin to selecting just a
few balls from the bag as representative of the entire contents of the
bag. In the case of VAEs, the compressed representation is constructed
with the goal of maximizing the fidelity of the reconstructions of the
input from the representation. It is an open question, however,
whether the compressed representations show any undue preference
for higher-level semantic information (the content of input images,
or example, as described by a human). 
If they do not, VAE compression is analogous to blindly
drawing balls from the bag, without preferentially selecting red balls
(analogous to higher-level semantic information). More compressed representations
would retain less semantic information, and the precise relationship might be
expected to be linear.
If they do, this should be detectable as a nonlinear relationship between
the semantic information retained and the compression rate
of the representation, and we might ask by which method
the VAE's are ``cheating'', somehow sensing the color (label information)
in an otherwise color-blind objective (unsupervised learning).

It was recently shown~\citep{brokenelbo,understandbeta}
that the $\beta$-VAE objective generally and the VAE objective specifically
can be understood to be variational approximations to a constrained
mutual information maximization objective:
\begin{equation}
	\max_{p_\theta(z|x)} I(Z;X) \text{ s.t. } I(Z;X) \leq I_0
\end{equation}
The goal is to learn a parametric representation of data that retains
salient information while compressing the data as much as possible.
This objective, however, does not make any distinction on its own as to
what constitutes salient information.  In fact, from the information
theoretic perspective, any information is salient.

Studies~\citep{landauer} suggest that when a human looks at a picture,
they extract only about a dozen bits into their long term memory.  The
class label of an MNIST digit, which a human immediately perceives,
can be encoded in 3.3 bits. A highly-compressed VAE encoding of a
115-bit binarized MNIST image may or may not contain the 3.3 bits
representing the digit. (If it does, it's akin to drawing
just a fraction of the total number of balls in the bag and pulling
out all 3 red ones.) This motivates our research question: to what
degree do our current approaches to unsupervised learning learn to
extract the same compressed set of bits that humans do?

}%

	\section{Methods}

To learn our representations, we use the $\beta$-VAE objective,
interpreted in a representational light~\citep{brokenelbo}.  The
encoder $p(z|x)$ learns a stochastic map from data $X$ (with density
$p(x)$) to some representation $Z$ that minimizes the objective $D +
\beta R$, where $D$ is the \emph{distortion}:
\begin{equation}
	D \equiv  -\mathbb{E}_{p(z,x)} \left[ \log d(x|z) \right] \geq H(X) - I(X;Z).
\end{equation}
The distortion is defined by the negative log likelihood of a trained
variational \emph{decoder} $d(x|z)$ that attempts to reconstruct the
image from the representation.  It provides, up to an additive
constant a lower bound on the mutual information between our image and
our representation.

The \emph{rate} ($R$) gives a variational upper bound on the mutual
information retained about the image:
\begin{equation}
	R \equiv \mathbb{E}_{p(z,x)} \left[ \log \frac{p(z|x)}{m(z)} \right] > I(Z;X),
\end{equation}
defined by means of the variational marginal $m(z)$ (usually called the prior for the
VAE when interpreted as a generative model). 
The rate measures the complexity of the learned representation, measured in
bits.  As an average KL divergence it denotes how many excess bits would be
required to communicate the encodings using an entropic code designed for the
variational marginal.

When $\beta=1$, the $\beta$-VAE objective is equivalent to the
Evidence Lower Bound (ELBO) of variational inference, in which rate is
the KL term and distortion is the likelihood term. See
\citet{brokenelbo} for a more thorough explanation of the connection
between $\beta$-VAEs and rate-distortion theory.

To measure the degree to which our representation has captured
\emph{salient} information, we measure the mutual information retained
about the \emph{labels}.  To do this, after training our $\beta$-VAE
we separately train a variational classifier to provide a lower bound
(up to an additive constant), of the mutual information between our
learned representation and the labels, the \emph{label distortion}
($C$).
\begin{equation}
	C \equiv -\mathbb{E}_{p(x,y,z)} \left[ \log c(y|z) \right] \geq H(Y) - I(Z;Y)
\end{equation}

We additionally train end-to-end classifiers, both of the simple two layer
fully connected network used above, as well as a convolutional network to
establish baseline supervised values for $C$.  We did a hyperparameter search
over filter depths and dropout rates to try to establish a good baseline.

In order to help distinguish how much the unsupervised learning could
benefit from gross geometric information contained in the data, we
compare to a simple stochastic PCA baseline.  We first project the
data by means of a truncated PCA down to a small number of dimensions
($k$).  From here we inject isotropic Gaussian noise of some magnitude
$\sigma$.  Assuming the data is now exactly Gaussian under the
whitening transformation, the mutual information between the input and
the representation can be computed exactly~\citep{mutinfonotes}:
\begin{equation}
	R_{\text{PCA}} = I(X; X + N) = \frac{k}{2} \log \left( 1 + \frac{1}{\sigma^2} \right).
\end{equation}

To test how much the VAEs, even if unsupervised might benefit from the
inductive biases introduced by the convolutional structure of the encoder, we
also measured the $C$ obtained from a randomly initialized two layer fully
connected neural network as the encoder, as well as a randomly initialized
convolutional network.  These networks were not trained; we simply used the
deterministic encoding produced by the randomly initialized architecture and
measured how much information they retained about the labels. Note that this is
an uncompressed setting, and so should be expected to retain most of the label
information.

We test the effect of different encoder architectures, decoder architectures, and marginal distributions on the degree of class-relevant information that the encodings capture.

}%

	\section{Experiments and discussion}
\label{sec:experiments}

The initial experimental setup is identical to \citet{brokenelbo},
where we trained many different architectures of $\beta$-VAEs for a
wide range of $\beta$s on the Binary Static MNIST dataset of
\citet{hugo} as well as Omniglot~\citep{omniglot}.  The labels for the
Binary Static MNIST dataset were obtained by naive Bayes with respect
to original MNIST.  We then used a two-layer fully-connected network
with ELU activations trained with Adam for 100 epochs with a learning
rate of 1e-4, decayed by a factor of 10 every 25 epochs.

For Omniglot we used the same data split as in \citet{iwae}.  This allowed us
to test both how much information was retained about the alphabet (a one in 50
classification task) as well as about the individual characters (a one in 1623
classification task).

The baseline convolutional classifier for Binary Static MNIST
had five 5x5 convolutional layers of depth 64 followed by a two
layer 200 unit fully connected classifier.  2x2 max pooling layers and
dropout (at a rate of 30\%) were used after layers 3 and 5.  For the
omniglot alphabet task, the same network was used without dropout and
deeper (256) convolutional filters. The omniglot character task used a
dropout rate of 10\% with depth 64 convolutional filters.

\begin{figure}[htbp]
	\centering
	\includegraphics[width=\textwidth]{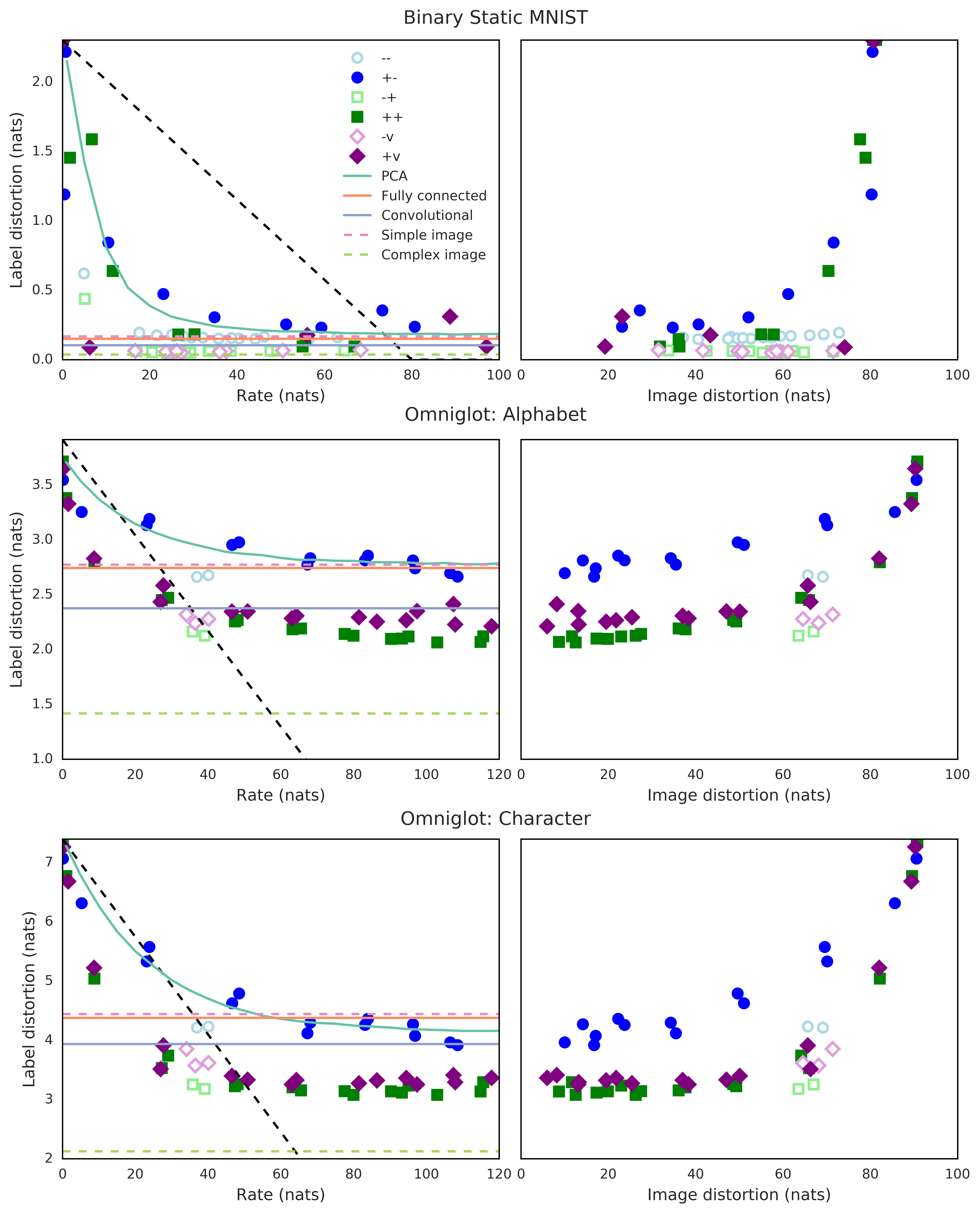}
	\caption{\label{fig:labelrate} Label distortions ($C$) versus
          rate ($R$) for all runs and baselines (left) and measured
          label distortions ($C$) versus image distortions ($D$) for
          all runs (right).  See text for explanation of baselines
          (lines).  The tuple in the key (d, m) denotes the relative
          power of the decoder (d) and marginal (m), in which $+$ is
          powerful, $-$ is simple, and $v$ is VampPrior.  Color and
          shape encodes the marginal, and fill denotes the decoder.  }
\end{figure}

\begin{figure}[htbp]
\centering
\includegraphics[width=\textwidth]{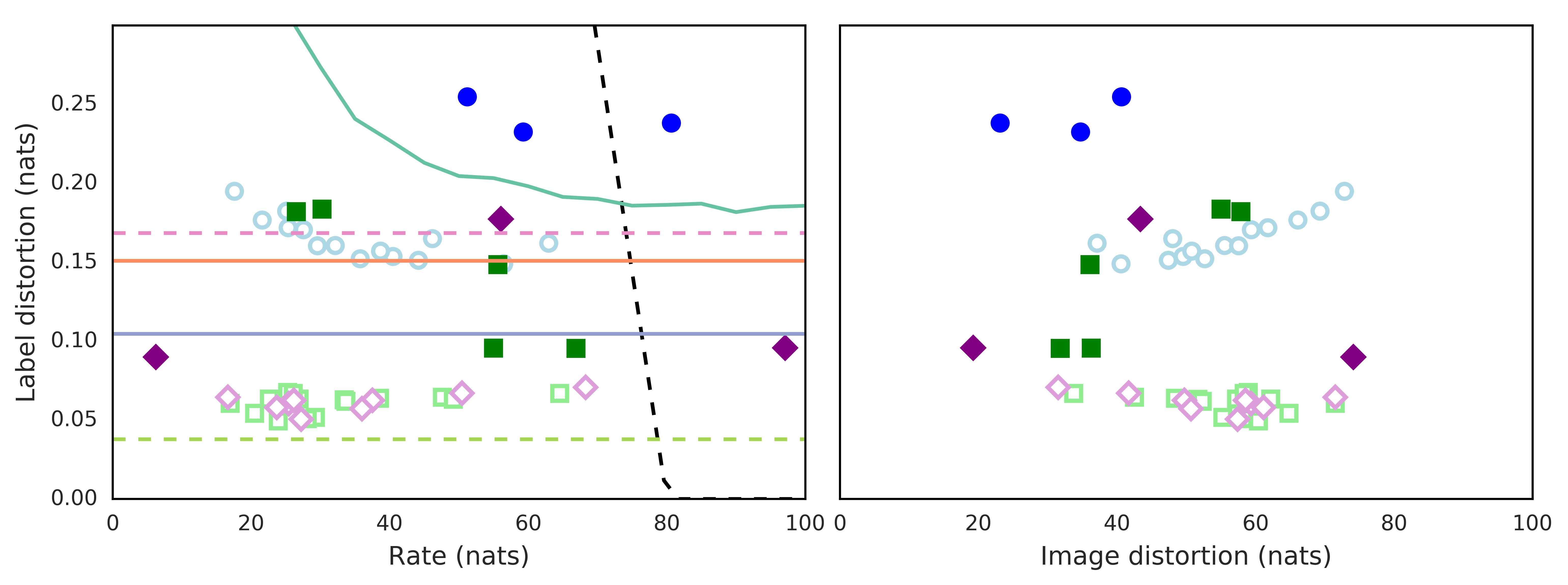}
\caption{\label{fig:mnistzoom} Label distortions versus rate (left)
  and image distortion (right) for binary static MNIST, magnified to
  show low label distortions (the data and legend are the same as the
  top subplots of Figure~\ref{fig:labelrate}).}
\end{figure}

\Cref{fig:labelrate} shows the main results.  The figures on the left
show the measured label distortions ($C$) versus rate for all runs and
baselines. This shows how much information was retained about the
labels (vertical, where lower is better) versus how compressed the
representation was (horizontal, left is more compressed).  The figures
on the right show the label distortion ($C$) versus the image
distortion ($D$).  This shows how much label information is retained
versus how well the VAEs were able to reconstruct the original images.
A zoomed in version of the MNIST images are shown in
\Cref{fig:mnistzoom}.

The dashed diagonal black line shows the performance of a method that
randomly discards information from an optimal lossless compression
(using an estimate of the entropy of 79.78 nats~\citep{vampprior}),
with no regard for the labels whatsoever. The solid green line
(labelled PCA) shows the best randomized PCA results for each dataset
(which were a dimensionality of 30 for MNIST and 42 for both Omniglot
tasks).  The solid red line (Fully connected) shows how much label
information was retained in the representation from a randomly
initialized fully connected network.  The solid blue line
(Convolutional) shows how much label information was retained from the
activations in a randomly initialized convolutional network. The
dashed red line (Simple image) shows the baseline fully trained
classifier performance working on the raw images, and lastly the
dashed green line (Complex image) shows the fully trained
convolutional network's label distortion.  The different network
architectures are shown with different symbols.

On MNIST, many runs consistently beat the simple baselines and achieve
very small label distortions ($C$, retain a large amount of label
information) at small rates ($R$, high compression), showing
preference for the label information.  Decoder architecture (whether
the symbol is solid or open) had a considerable impact on what
information the encoder captured: lower label distortions were
achieved when a weaker factorized decoder was used rather than the
more powerful auto-regressive variant.  This is likely because when
the decoder is factorized, the mutual information between the latent
and each factor has an independent additive effect on the
distortion. For a factorized decoder, distortion (up to an additive
constant) is:
\begin{equation}
  D \equiv \sum_i KL(p(x_i|z)||d(x_i|z)) - \sum_i I(X_i; Z)
\end{equation}
In the above equation the $KL$ term measures the gap between the
conditional likelihood of each factor
($p(x_i|z)=\frac{p(x_i,z)}{p(z)}$) and its variational approximation
(i.e the decoder) and the other term ($\sum_i I(X_i;Z)$) adds the
mutual information with the latent for each factor.  Consequently, in
a low rate setting, the encoder can be biased to capture information
that is shared and manifested in more factors (pixels), which our
experiments indicate has led to the encoding of more salient features
of an image.

On MNIST, (\Cref{fig:mnistzoom}) as early as 8 nats in a single
instance (which we note used an autoregressive decoder) and around 17
nats generically (for deconvolutional decoders), the $\beta$-VAEs,
were retaining a large amount of label information.  They clearly show
some preference for label information (i.e. they are much better than
the dashed black baseline).  Their preference is stronger than can be
explained by PCA (solid green line).  The inductive biases introduced
by convolutional networks seem to be important but we emphasize that
the $\beta$-VAEs were retaining more label information than the random
\emph{uncompressed} convolutional network and within 0.05 nats (0.07
bits) of the most label information able to be extracted even
by a fully trained \emph{uncompressed} network.  These baselines
are of a different kind than the $\beta$-VAEs, having unbounded rate.

These $\beta$-VAE networks, even though there were trained in a
completely unsupervised manner and were forced to compress the inputs
by a factor of 4 (at a rate near 20), retained nearly all
of the label information.  In one case, an MNIST $\beta$-VAE with an
autoregressive decoder retained nearly all of the label information at
a compression factor over 9.

On the Omniglot Alphabet classification task (second row of
\Cref{fig:labelrate}), the encodings retain much less of the label
information.  Most of the $\beta$-VAE points don't show evidence for
preferring the alphabet information (most points are near or to the
right of the black dashed line). 
It is also interesting to note that the autoregressive
decoder architectures with the traditional choice of a fixed isotropic
Gaussian marginal (prior, solid blue circles) track the randomized PCA
baseline.  
The effect of the prior appears
greater for Omniglot than for MNIST, indicating that the Gaussian prior
was less able to capture informative encodings for the richer Omniglot
dataset. 
There are a few networks at a rate near 10 that show some
preference for alphabet class information, but the effect is much
weaker.  The $\beta$-VAEs show more preference for the character
identities (third row of \Cref{fig:labelrate}), but the effect, if
still present, is much weaker than on MNIST. This suggests trying much
more powerful models on Omniglot to see if the effect can be restored.

\section{Conclusion}

In conclusion, we have shown that the encodings of a $\beta$-VAE can capture a
significant amount of information that is salient to the class labels,
even at high compression rates.  This should be surprising given that
$\beta$-VAEs are trained in a completely unsupervised way. 
The effect is a strong function of decoder architecture, and dataset.

In future work, we plan to test the effect of alternate lower bounds
for $I(X, Z)$ and $I(Z, Y)$, such as those recently proposed in
\citet{cpc,mine}. Corroborating our results across multiple lower
bounds will strengthen conclusions on the effects of model
architecture on $I(Z, Y)$. In addition to VAEs, we intend to
investigate the extent to which latent representations of other
generative models (such as normalizing flows) retain class
information. Using the proposed method on richer multiclass datasets,
such as CelebA, will enable us to further examine the characteristics
of the class labels that are retained in the encodings of different
models.
}%

        \bibliography{bib}

\begin{thebibliography}{10}
\providecommand{\natexlab}[1]{#1}
\providecommand{\url}[1]{\texttt{#1}}
\expandafter\ifx\csname urlstyle\endcsname\relax
  \providecommand{\doi}[1]{doi: #1}\else
  \providecommand{\doi}{doi: \begingroup \urlstyle{rm}\Url}\fi

\bibitem[Alemi et~al.(2018)Alemi, Poole, Fischer, Dillon, A., and
  Murphy]{brokenelbo}
Alexander~A. Alemi, Ben Poole, Ian Fischer, Joshua~V. Dillon, Saurous~Rif A.,
  and Kevin Murphy.
\newblock Fixing a broken elbo.
\newblock \emph{ICML}, 2018.
\newblock URL \url{https://arxiv.org/abs/1711.00464}.

\bibitem[Belghazi et~al.(2018)Belghazi, Rajeswar, Baratin, Hjelm, and
  Courville]{mine}
Ishmael Belghazi, Sai Rajeswar, Aristide Baratin, R~Devon Hjelm, and Aaron
  Courville.
\newblock Mine: mutual information neural estimation.
\newblock \emph{ICML}, 2018.
\newblock URL \url{https://arxiv.org/abs/1801.04602}.

\bibitem[Burda et~al.(2015)Burda, Grosse, and Salakhutdinov]{iwae}
Yuri Burda, Roger Grosse, and Ruslan Salakhutdinov.
\newblock Importance weighted autoencoders.
\newblock 2015.

\bibitem[{Burgess} et~al.(2018){Burgess}, {Higgins}, {Pal}, {Matthey},
  {Watters}, {Desjardins}, and {Lerchner}]{understandbeta}
C.~P. {Burgess}, I.~{Higgins}, A.~{Pal}, L.~{Matthey}, N.~{Watters},
  G.~{Desjardins}, and A.~{Lerchner}.
\newblock {Understanding disentangling in $\beta$-VAE}.
\newblock \emph{arXiv}, 2018.
\newblock URL \url{https://arxiv.org/abs/1804.03599}.

\bibitem[Lake et~al.(2015)Lake, Salakhutdinov, and Tenenbaum]{omniglot}
Brenden~M. Lake, Ruslan Salakhutdinov, and Joshua~B. Tenenbaum.
\newblock Human-level concept learning through probabilistic program induction.
\newblock \emph{Science}, 350\penalty0 (6266):\penalty0 1332--1338, 2015.

\bibitem[Landauer(1986)]{landauer}
Thomas~K Landauer.
\newblock How much do people remember? some estimates of the quantity of
  learned information in long-term memory.
\newblock \emph{Cognitive Science}, 10\penalty0 (4):\penalty0 477--493, 1986.

\bibitem[Larochelle \& Murray(2011)Larochelle and Murray]{hugo}
Hugo Larochelle and Iain Murray.
\newblock The neural autoregressive distribution estimator.
\newblock 2011.

\bibitem[Polyanskiy \& Wu(2017)Polyanskiy and Wu]{mutinfonotes}
Y~Polyanskiy and Y~Wu.
\newblock Lecture notes on information theory, 2017.
\newblock URL
  \url{http://people.lids.mit.edu/yp/homepage/data/itlectures_v5.pdf}.

\bibitem[{Tomczak} \& {Welling}(2017){Tomczak} and {Welling}]{vampprior}
J.~M. {Tomczak} and M.~{Welling}.
\newblock {VAE with a VampPrior}.
\newblock \emph{ArXiv e-prints}, 2017.

\bibitem[{van den Oord} et~al.(2018){van den Oord}, {Li}, and {Vinyals}]{cpc}
A.~{van den Oord}, Y.~{Li}, and O.~{Vinyals}.
\newblock {Representation Learning with Contrastive Predictive Coding}.
\newblock \emph{arXiv:1807.03748}, July 2018.
\newblock URL \url{https://arxiv.org/abs/1807.03748}.

\end{thebibliography}
        \bibliographystyle{iclr2019_conference}

\end{document}